\documentclass[sigconf]{acmart}
\AtBeginDocument{%
  }

\setcopyright{acmlicensed}
\copyrightyear{2026}
\acmYear{2026}
\acmDOI{XXXXXXX.XXXXXXX}
\acmConference[MM'26]{the 34th ACM International Conference on Multimedia}{November 10--14, 2026}{Rio de Janeiro, Brazil}
\acmISBN{978-1-4503-XXXX-X/2018/06}

\begin{document}

\title{Decoupling Cross-Modality Manifold Discrepancy: Leveraging Visible Diffusion Priors for Infrared Super-Resolution}

\author{Yunpeng Hua}
\authornote{Equal contribution.}
\affiliation{
	\institution{University of Science and Technology Beijing}
	\city{Beijing}
	\country{China}
}
\email{huayunpeng2011@126.com}

\author{Hongwei Yu}
\authornotemark[1]
\affiliation{
  \institution{University of Science and Technology Beijing}
  \city{Beijing}
  \country{China}
}
\email{yuhongwei22@xs.ustb.edu.cn}

\author{Jiawei Li}
\affiliation{
  \institution{University of Science and Technology Beijing}
  \city{Beijing}
  \country{China}
}
\email{ljw19970218@163.com}

\author{Qiankun Liu}
\affiliation{
  \institution{University of Science and Technology Beijing}
  \city{Beijing}
  \country{China}
}
\email{liuqk3@ustb.edu.cn}

\author{Huimin Ma}
\affiliation{
	\institution{University of Science and Technology Beijing}
	\city{Beijing}
	\country{China}
}
\email{mhmpub@ustb.edu.cn}

\author{Jiansheng Chen}
\authornote{Corresponding author.}
\affiliation{
	\institution{University of Science and Technology Beijing}
	\city{Beijing}
	\country{China}
}
\email{jschen@ustb.edu.cn}

\renewcommand{\shortauthors}{Yunpeng Hua et al.}

\begin{abstract}
Infrared image super-resolution (IISR) mitigates the limitations imposed by low spatial resolution. Existing methods have recognized that IISR should preserve consistency in global distribution and structural information while enhancing image clarity. However, these methods are either insufficient or overly intrusive, a problem that becomes even more pronounced in diffusion-based models. To address these issues, we propose a dual-path diffusion-based framework for IISR, termed Shift-IISR. The proposed method is designed to improve the consistency of IISR results while preserving the generative capacity of diffusion models. Specifically, we develop a Global Representation Modulation (GRM) module to extract modality-specific information from infrared imagery and guide the global distribution of the diffusion model toward the ground truth. In addition, we introduce a Local Structure Refinement (LSR) module to encourage the model to focus on structural information at each step of the iterative denoising process. Extensive experiments demonstrate that the proposed method effectively improves distributional and structural consistency while maintaining competitive super-resolution performance. The source code of the proposed Shift-IISR can be available at https://github.com/Assassink8/Shift-IISR.

\end{abstract}


\begin{teaserfigure}
  \includegraphics[width=\textwidth]{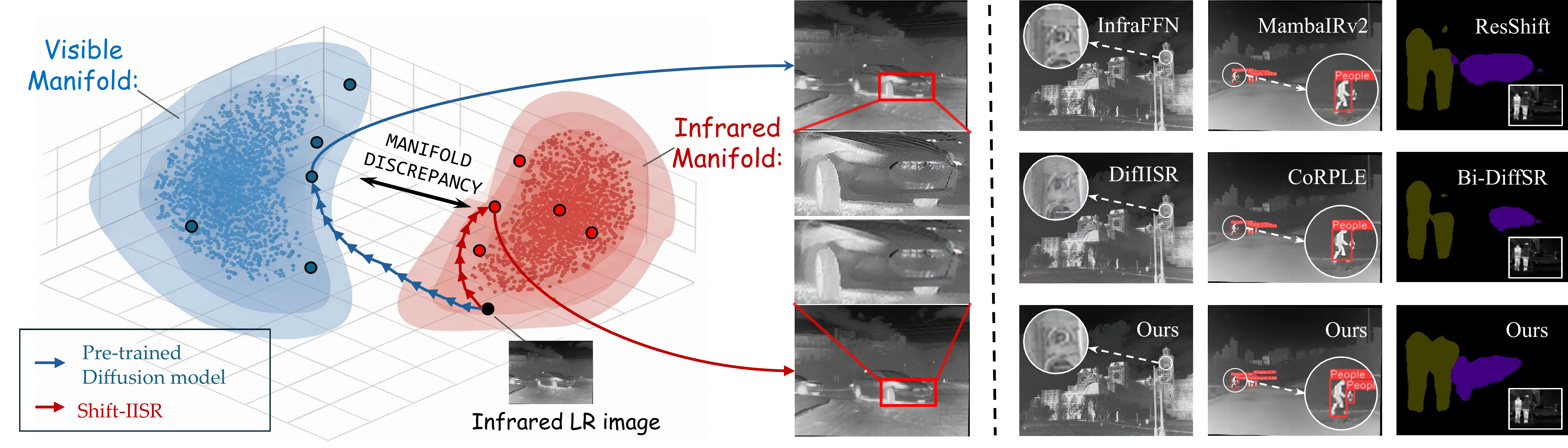}
  \caption{The left illustrates the limitations of a pre-trained diffusion model for infrared image super-resolution. Shift-IISR effectively guides the model toward the infrared manifold. The right demonstrates the superior performance of our method in terms of visual quality and downstream tasks.}
  \Description{manifold discrepancy}
  \label{fig:teaser}
\end{teaserfigure}


\maketitle

\section{Introduction}
Infrared imagery provides robust thermal information and remains effective under low- or no-light conditions \cite{li2023gesenet}, making it highly valuable for applications such as embodied intelligence \cite{zhong2025darkseg, li2026multi}, autonomous driving \cite{cao2023multi}, and object detection \cite{liu2022target,li2025a2rnet,li2023learning}. However, constrained by hardware limitations, infrared images inherently suffer from low resolution, poor contrast, and blurring \cite{liu2025enhancing}. To address these issues, Infrared Image Super-Resolution (IISR) has become a focal point of research \cite{wang2020deep}. Unlike visible image super-resolution, however, the goal of IISR is not merely to recover visually pleasing details, but to enhance spatial resolution while preserving the modality-specific characteristics rooted in thermal imaging. The mismatch between the task objective and the technical paradigm suggests that IISR should prioritize consistency in global distribution and local structure, rather than natural textures and sharp edges that may cause distortion \cite{ma2019infrared,kawar2022denoising}.

As research has progressed, increasing attention has been devoted to the unique characteristics of IISR, and various efforts have been made to improve its consistency. Some studies enhance global distribution consistency through feature alignment or fine-tuning strategies, while others introduce edge constraints to strengthen structural perception. More recently, diffusion models \cite{saharia2022image, li2022srdiff, shang2024resdiff, luo2023image} have emerged as a popular paradigm for IISR, leading to the development of various diffusion frameworks through the exploration of conditional constraints and improved efficiency.

Although the above studies have attempted to bridge the mismatch between technical paradigms and the objectives of IISR through different strategies, several limitations remain. (i) Owing to the scarcity of infrared data, it is challenging to build a high-quality diffusion framework for IISR from scratch. Meanwhile, transfer-based strategies, such as fine-tuning pre-trained models on limited infrared data, can easily compromise the original generative capability of the model \cite{moser2024diffusion}. (ii) Feature alignment and related strategies for improving global distribution consistency cannot fundamentally correct the visible-light priors embedded in the model \cite{isola2017image, li2020multigrained}. (iii) Enhancing structural fidelity via edge-based constraints is feasible \cite{saharia2022image, ma2020structure}, but such static regularization is unable to precisely control the diffusion trajectory in diffusion models characterized by progressive iterative reconstruction. 

To address the above limitations, we propose a dual-path diffusion-based framework for IISR, termed Shift-IISR. In response to the difficulty of training a high-quality infrared diffusion model from scratch and the risk that fine-tuning may impair the generative capability of pre-trained models, our method keeps the diffusion backbone frozen, thereby preserving its original generative prior, while leveraging two dedicated modules to improve global distribution consistency and local structural consistency for IISR. We argue that the distribution inconsistency in IISR mainly stems from the visible-biased prior embedded in the pre-trained diffusion model. To address this, we introduce a Global Representation Modulation (GRM) module to learn the relative modality shift under the reference of same-scene visible representations. By contrasting paired infrared and visible latent features, GRM extracts infrared-specific distributional characteristics and injects the learned infrared prior into the denoising process through time-embedding modulation, thereby progressively correcting the visible-biased generative prior and improving global distribution consistency. Meanwhile, a Local Structure Refinement (LSR) module is proposed to overcome the limitation of static edge-based constraints in controlling the progressive reconstruction trajectory. It injects structural information into the denoised result at each time step and progressively guides the sampling process toward structurally faithful reconstructions. A predefined time-step-dependent function is similarly introduced in LSR to control the refinement strength across different stages. Through this dual-path design, the proposed framework retains most of the original generative capability of the pre-trained model while effectively improving both global distribution consistency and local structural fidelity, thereby progressively steering the denoising trajectory back to the intrinsic infrared manifold, as illustrated in Fig. \ref{fig:teaser}. Extensive experiments demonstrate that our method not only achieves competitive super-resolution performance, but also effectively enhances both global distributional consistency and structural consistency, leading to reconstructions that are more faithful to the ground truth in terms of both overall intensity distribution and local structural details. Our main contributions are summarized as follows:
\begin{itemize}
	\item We propose a novel dual-path diffusion-based framework for infrared image super-resolution (IISR), which improves both global distribution and local structural consistency while largely retaining the generative capabilities of the pre-trained diffusion model.
	
	\item We introduce a Global Representation Modulation (GRM) module that progressively guides the denoising process through feature injection, thereby suppressing the impact of visible-light priors and improving distributional consistency with the ground truth to some extent.
	
	\item We propose a Local Structure Refinement module that integrates structural cues into each denoising step, effectively suppressing structural artifacts and enhancing the geometric fidelity of the reconstructed results.
	
	\item Extensive experiments demonstrate the effectiveness of the proposed method, showing that it not only achieves competitive super-resolution performance but also significantly improves the global distributional and structural consistency of infrared images.
\end{itemize}

\section{Related work}
\subsection{Image Super-Resolution}

The field of image super-resolution has advanced remarkably over the past decade, evolving from reconstruction-based methods to more powerful generative paradigms. Early progress was largely driven by Convolutional Neural Networks (CNNs) \cite{jiang2024multispectral, luo2020latticenet, zhang2015ccr, zhang2018image, zhang2018residual}, with representative models such as SRCNN \cite{dong2014learning}, EDSR \cite{lim2017enhanced}, and RCAN \cite{zhang2018image} significantly improving pixel-wise reconstruction accuracy. To mitigate the over-smoothing effects of MSE-based optimization, Generative Adversarial Networks (GANs) \cite{zhang2021designing, wang2021real, kang2023scaling, chen2023activating}, including SRGAN \cite{ledig2017photo} and ESRGAN \cite{wang2018esrgan}, were introduced to enhance perceptual realism through high-frequency detail synthesis. More recently, Vision Transformers (ViTs) \cite{cao2021video, chen2021pre, zamir2022restormer, zhang2022efficient} have further improved structural reconstruction by modeling long-range dependencies, as demonstrated by methods such as SwinIR \cite{liang2021swinir} and HAT \cite{chen2023activating}. 

Despite these breakthroughs in the visible domain, research specifically tailored for IISR remains relatively sparse. CoRPLE \cite{li2024contourlet} introduces contourlet-based priors to enhance structural representation, DifIISR \cite{li2025difiisr} adapts the pre-trained ResShift framework to the infrared domain, and InfraFFN \cite{qin2025infraffn} improves reconstruction through frequency-aware feature modeling. Despite these advances, existing methods still struggle to properly preserve the intrinsic characteristics of infrared imagery during reconstruction.

\begin{figure*}[t]
	\centering
	\includegraphics[width=\textwidth]{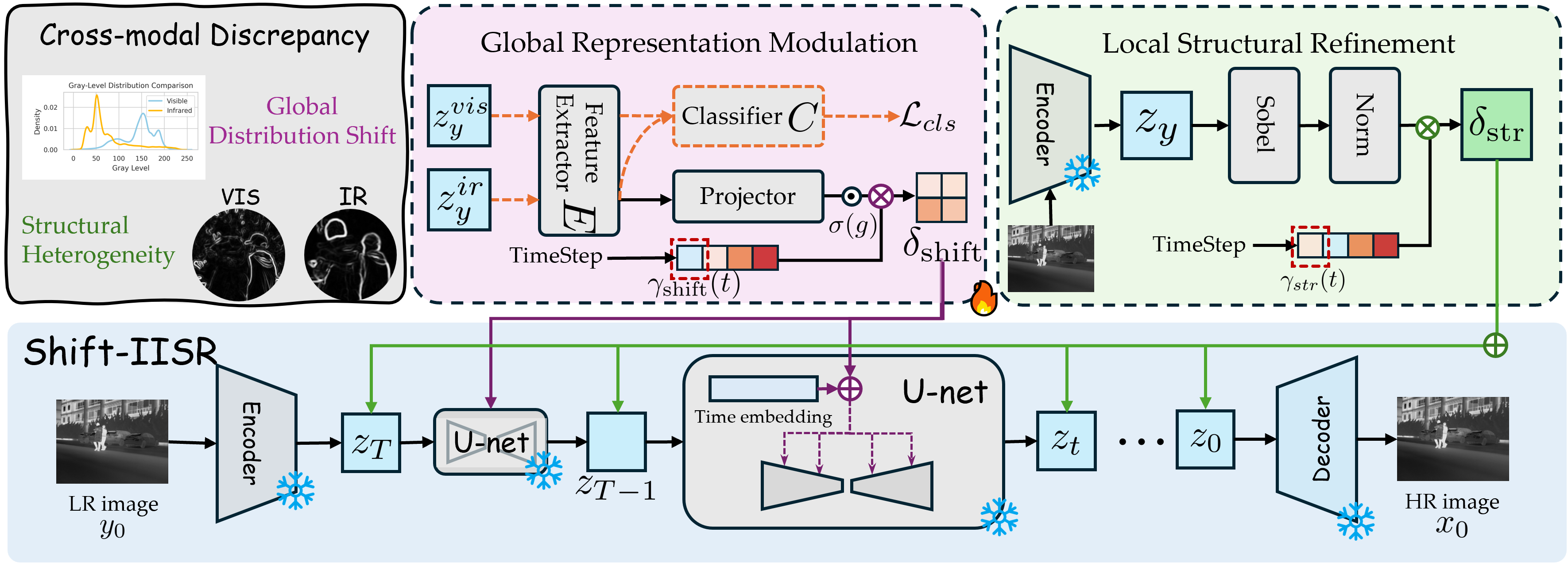}
	\caption{Overall architecture of Shift-IISR: cross-modal discrepancies between visible
		and infrared domains are addressed by guiding the diffusion model via GRM
		(\textcolor[HTML]{4B0082}{Purple}) and LSR
		(\textcolor[HTML]{006400}{Green}), enabling IISR. The dashed paths
		\textcolor[HTML]{E97132}{(Orange)} indicate training-only operations.}
	\label{fig:model}
\end{figure*}

\subsection{Diffusion Models}
In recent years, image restoration has been significantly advanced by diffusion models, which are highly effective at modeling complex data distributions. Denoising Diffusion Probabilistic Models (DDPM) \cite{ho2020denoising} established the basic framework of stochastic iterative refinement and demonstrated strong capability in generating high-fidelity textures. To improve sampling efficiency, Denoising Diffusion Implicit Models (DDIM) \cite{song2020denoising} introduced non-Markovian sampling for faster inference, while Latent Diffusion Models (LDM) \cite{rombach2022high} further reduced computational cost by performing diffusion in a compressed latent space without sacrificing generation quality.

To better exploit diffusion models for image restoration, various conditioning strategies have been developed to enhance spatial and semantic controllability, including input-level concatenation \cite{saharia2022palette}, conditional embedding modulation \cite{zhang2023adding}, context feature injection \cite{xia2023diffir}, and iterative residual correction \cite{shi2024resfusion}. These methods have achieved strong performance in blind SR and realistic image synthesis \cite{lin2024diffbir, wang2024exploiting}. However, extending such success to infrared image super-resolution remains challenging, since standard diffusion models depend on large-scale diverse training data, whereas high-quality infrared pairs are highly limited. As a result, adapting pre-trained generative priors to infrared-specific characteristics under data-scarce settings remains a challenging and under-explored problem.

\section{Problem Formulation}
\label{sec:pf}

Our method is built upon ResShift~\cite{yue2023resshift}, which replaces the conventional noise-to-data diffusion trajectory with a residual shifting process. Rather than progressively corrupting an HR image into isotropic Gaussian noise, ResShift gradually shifts the HR target toward its LR condition while injecting controlled Gaussian noise. For computational efficiency, this process is performed in the latent space of a pre-trained VQGAN~\cite{esser2021taming}.

Let $x_0$ and $y_0$ denote the HR target and LR input, with corresponding latent representations $z_0$ and $z_y$, respectively. Defining the residual shift as $e_0=z_y-z_0$, the forward process follows a monotonically increasing schedule $\alpha_t$:
\begin{equation}
	q(z_t \mid z_0,z_y)
	=
	\mathcal{N}\left(
	z_t;\,
	(1-\alpha_t)z_0+\alpha_t z_y,\,
	\kappa^2\alpha_t\mathbf{I}
	\right),
\end{equation}
where $\kappa$ controls the noise scale. Equivalently, the latent state at timestep $t$ can be expressed as
\begin{equation}
	z_t
	=
	z_0+\alpha_t e_0
	+
	\kappa\sqrt{\alpha_t}\,\epsilon,
	\qquad
	\epsilon\sim\mathcal{N}(\mathbf{0},\mathbf{I}).
\end{equation}
Thus, increasing $t$ progressively moves the latent from the HR target toward the LR condition.

Different from conventional noise-prediction diffusion models, ResShift directly predicts the clean HR latent from the shifted latent $z_t$, conditioned on $z_y$. The training objective is
\begin{equation}
	\label{eq:base-loss}
	\mathcal{L}_{\mathrm{simple}}
	=
	\mathbb{E}_{z_0,z_y,\epsilon,t}
	\left[
	\left\|
	z_0-f_\theta(z_t,z_y,t)
	\right\|^2
	\right].
\end{equation}
At inference, the predicted clean latent is iteratively refined through the reverse Markov chain defined by ResShift.

However, the visible-spectrum prior encoded in the pre-trained diffusion model may bias the predicted HR latent away from the infrared domain. Since reverse sampling repeatedly relies on these predictions, such cross-modality errors can accumulate along the reconstruction trajectory. These observations motivate a framework that jointly regulates global cross-modality alignment and local structural fidelity during reverse sampling.

\section{Method}
\subsection{Overview}
Building upon the latent-space ResShift framework, we introduce a Global Representation Modulation (GRM) module to provide auxiliary infrared specific representations to the denoising network at each sampling step. This global information is seamlessly integrated into the time embeddings, thereby facilitating continuous guidance across the entire iterative process. To address the challenge of structural heterogeneity, we develop a Local Structural Refinement (LSR) module that extracts salient edge priors and directly incorporates them into the intermediate sampling states at each iteration. By imposing these explicit geometric constraints, the module effectively guides the structural synthesis during the reverse process. Accordingly, we refine the estimation of the clean high-resolution latent during reverse sampling. Let $\delta_{\mathrm{shift}}$ denote the global distribution adjustment extracted by the GRM module and $\delta_{\mathrm{str}}$ represent the geometric prior from the LSR module. The modified single-step sampling rule is formulated as follows:

\begin{align}
	z_{t-1}
	&=
	\frac{\eta_t}{\alpha_t}
	f_\theta\!\bigl(
	\tilde z_t, z_y, \tau(t,\delta_{\mathrm{shift}})
	\bigr)
	+
	\frac{\alpha_{t-1}}{\alpha_t}\tilde z_t \nonumber \\
	&\quad +
	\kappa \sqrt{\frac{\alpha_{t-1}\eta_t}{\alpha_t}}\,\epsilon',
	\qquad
	\epsilon' \sim \mathcal{N}(0,\mathbf{I}).
	\label{all}
\end{align}
where $\tilde z_t=z_t+\alpha_t\delta_{\mathrm{str}}(t)$ denotes the structure-guided latent at timestep $t$, and $\tau(\cdot)$ represents the modulation function that injects infrared distribution cues into the temporal embedding. This formulation corrects the global statistical shift within the model's internal feature space, while injecting local structural guidance into the latent before reverse sampling, collectively steering the synthesis toward the intrinsic infrared manifold, as illustrated in Fig.~\ref{fig:model}.

\subsection{Global Representation Modulation}

\begin{figure}[h]
	\centering
	\includegraphics[width=\linewidth]{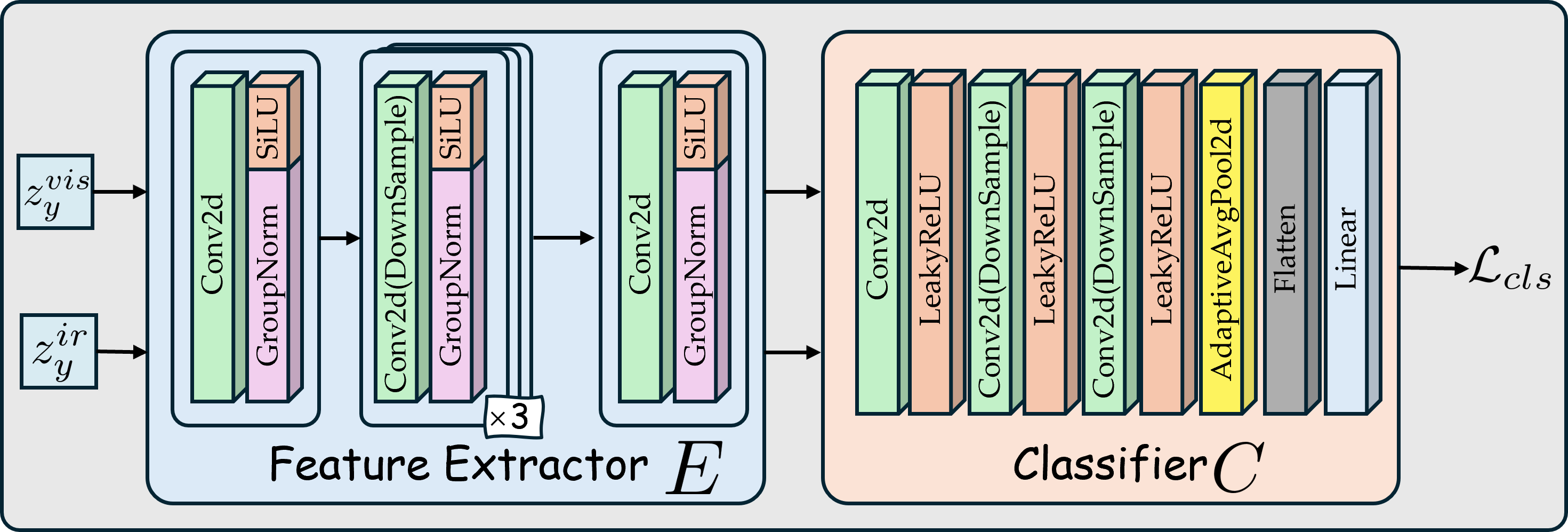}
	\caption{he architectures of the feature extractor $E$ and classifier $C$ in the GRM module.}
	\label{fig:GRM}
\end{figure}

To capture domain-specific infrared characteristics within the latent space, we introduce the Global Representation Modulation (GRM) module. Unlike conventional methods relying on explicit statistical indicators, GRM utilizes latent representations to implicitly model cross-modal discrepancies. As illustrated in Fig. \ref{fig:GRM}, the module employs a weight-sharing extractor $E$ to derive high-level semantic and stylistic features from the infrared domain.

Specifically, $E$ is implemented as a lightweight convolutional encoder designed to distill multi-scale statistical characteristics from the latent space. The encoder comprises an initial convolutional layer followed by $N$ successive downsampling stages ($N=3$ in our implementation). Each stage consists of a stride-2 convolution, Group Normalization (GN), and a Sigmoid Linear Unit (SiLU) activation function. This hierarchical downsampling expands the receptive field and reduces spatial resolution to encapsulate global modality-specific distributions. Ultimately, these features represent the modality-specific global representations $f_{ir}$ and $f_{vis}$ for infrared and visible images, respectively:

\begin{equation}
	f_{ir} = E(z_y^{ir}), \quad f_{vis} = E(z_y^{vis}),
\end{equation}
where $z_y^{ir}$ and $z_y^{vis}$ denote the latent space features of the low-resolution (LR) infrared and visible image extracted by the encoder, respectively.

To model relative modality shifts, we utilize same-scene visible latent features $z_y^{vis}$ as a training benchmark. By contrasting these modalities, the extractor $E$ learns to isolate unique infrared textural and distributional features. For robust supervision, we design a convolution-based classifier $C$ comprising sequential convolutions, non-linear activations, and adaptive pooling. Specifically, $C$ employs strided convolutions to capture spatial correlations and reduce dimensionality, outputting prediction probabilities via global pooling and a fully connected layer. Finally, using binary labels (0 for IR, 1 for visible), $C$ is optimized via cross-entropy loss:

\begin{equation}
	\label{cls}
	\mathcal{L}_{cls} = - \mathbb{E} \left[ \log(1 - C(f_{ir})) + \log(C(f_{vis})) \right].
\end{equation}
$\mathcal{L}_{cls}$ enforces the feature extractor $E$ to yield highly discriminative global representations, enabling $f_{ir}$ to accurately characterize the statistical discrepancies between infrared and visible spectra. Following dimensional alignment via a linear projector, a learnable gating parameter $g$, initialized to -2.0, is introduced to facilitate the seamless integration of the infrared prior. 

By applying a Sigmoid function $\sigma(\cdot)$ to adaptively modulate the injection intensity, the final infrared prior $\delta_{\mathrm{shift}}$ is derived for the generative sampling process:

\begin{equation}
	\delta_{\mathrm{shift}} = \sigma(g) \cdot \mathrm{Proj}(f_{ir}).
\end{equation}
For effective integration into the generative trajectory, the infrared prior is fused with the temporal embedding $e_t$ within the ResShift UNet blocks. We further employ a time-varying evolution schedule $\gamma_{shift}(t)$, which decreases monotonically with the forward timestep $t$ and complements the LSR schedule, to regulate the prior's influence across distinct diffusion stages, ensuring adaptive sensitivity to modal information:
\begin{equation}
	\tau(t, \delta_{\mathrm{shift}})=e_t+\gamma_{\mathrm{shift}}(t)\delta_{\mathrm{shift}}.
\end{equation}

We bridge domain identification and generative enhancement via a modal classifier co-trained with the GRM branch in an alternating manner within each iteration. In one sub-step, the classifier is optimized with $\mathcal{L}_{cls}$ using detached GRM features to establish a stable decision boundary; in the other, with the classifier frozen, its classification loss serves as an auxiliary objective alongside the denoising loss for optimizing the GRM branch, From Eqs. \ref{eq:base-loss} and \ref{cls}, the total generative loss $\mathcal{L}_{total}$ can be derived as:
\begin{equation}
	\mathcal{L}_{total} = \mathcal{L}_{simple} + \lambda \mathcal{L}_{cls}.
\end{equation}
$\lambda$ balances generative fidelity against modal constraint intensity. This configuration compels $E$ to characterize modal discrepancies while ensuring representation compatibility with the diffusion backbone. Additionally, this alternating strategy bolsters convergence stability and adaptation efficiency.

\begin{table*}[t]
	\centering
	\caption{Quantitative comparison on Set5, Set15, and Set20. Best results are in bold, and second-best results are underlined.}
	\label{tab:quantitative_comparison}
	\begin{tabular*}{\textwidth}{@{\extracolsep{\fill}}llccccccccc@{}}
		\toprule
		\multicolumn{2}{c}{Datasets} 
		& \multicolumn{3}{c}{\textbf{Set5}} 
		& \multicolumn{3}{c}{\textbf{Set15}} 
		& \multicolumn{3}{c}{\textbf{Set20}} \\
		
		\cmidrule(lr){1-2} \cmidrule(lr){3-5} \cmidrule(lr){6-8} \cmidrule(lr){9-11}
		
		\multicolumn{2}{c}{Methods} 
		& PSNR$\uparrow$ & LPIPS$\downarrow$ & SSIM$\uparrow$
		& PSNR$\uparrow$ & LPIPS$\downarrow$ & SSIM$\uparrow$
		& PSNR$\uparrow$ & LPIPS$\downarrow$ & SSIM$\uparrow$\\
		
		\midrule
		ResShift \cite{yue2023resshift}   & NeurIPS'23 & 28.360 & 0.490 & 0.648 & 27.716 & 0.414 & 0.716 & 28.726 & 0.436 & 0.699 \\
		SinSR \cite{wang2024sinsr}      & CVPR'24    & 32.400 & 0.306 & 0.851 & 31.501 & 0.270 & 0.865 & 32.966 & 0.244 & 0.888 \\
		SinSRv2 \cite{wang2024sinsr}      & CVPR'24    & 32.679 & 0.291 & 0.875 & 31.333 & 0.276 & 0.864 & 32.940 & 0.255 & 0.891 \\
		BI-DiffSR \cite{chen2024binarized}  & NeurIPS'24 & 30.126 & 0.364 & 0.783 & 28.558 & 0.352 & 0.786 & 29.961 & 0.364 & 0.790 \\
		ATD \cite{zhang2024transcending}        & CVPR'24    & 31.004 & 0.262 & 0.903 & 29.611 & 0.238 & 0.893 & 31.390 & 0.249 & 0.915 \\
		MambaIR \cite{guo2024mambair}    & ECCV'24    & 31.493 & 0.253 & 0.906 & 30.146 & 0.226 & 0.900 & 31.790 & 0.242 & 0.919 \\
		MambaIRv2 \cite{guo2025mambairv2}  & CVPR'25    & 31.950 & 0.239 & 0.914 & 30.389 & 0.219 & 0.903 & 31.904 & 0.239 & 0.920 \\
		\midrule
		CoRPLE \cite{li2024contourlet}     & ECCV'24    & \textbf{35.269} & 0.215 & \underline{0.934} & \textbf{33.441} & \underline{0.195} & \underline{0.924} & \textbf{34.555} & 0.239 & \underline{0.933} \\
		InfraFFN \cite{qin2025infraffn}   & KBS'25     & 33.708 & \underline{0.214} & 0.924 & 31.937 & 0.198 & 0.912 & 33.064 & \underline{0.236} & 0.922 \\
		DifIISR \cite{li2025difiisr}    & CVPR'25    & 32.396 & 0.317 & 0.864 & 31.359 & 0.289 & 0.872 & 32.898 & 0.268 & 0.895 \\
		\midrule
		Shift-IISR  & Ours       & \underline{35.129} & \textbf{0.162} & \textbf{0.935} & \underline{33.375} & \textbf{0.156} & \textbf{0.927} & \underline{34.384} & \textbf{0.191} & \textbf{0.934} \\
		\bottomrule
	\end{tabular*}
\end{table*}

\subsection{Local Structure Refinement}
While the GRM module provides guidance for the global statistical shift, the preservation of local structural integrity remains equally indispensable for super-resolution tasks. To this end, we propose the Local Structural Refinement (LSR) module to preserve structural integrity by injecting local spatial priors into the denoising process. By leveraging explicit edge priors, the LSR module facilitates the reconstruction of sharp, well-defined boundaries, ensuring that local geometric details are faithfully restored.

As depicted in Fig. \ref{fig:model}, the LSR module initially employs a Sobel operator on the low-resolution latent features $z_y$ to characterize local gradient variations by computing the weighted differences in both horizontal and vertical directions. To maintain computational stability and mitigate the influence of inherent contrast discrepancies, we perform mean-normalization on the extracted edge features, thereby deriving the refined structural feature $\delta_{norm}$:

\begin{equation}
	\delta_{norm} = \frac{S(z_y)}{\text{mean}(|S(z_y)|) + \xi} \, ,
\end{equation}
where $\xi$ denotes a small positive constant, set to $1 \times 10^{-6}$ in our experiments, to prevent division by zero and ensure numerical stability during the normalization process.

Prior work \cite{park2023understanding} indicates that structural guidance requirements vary across the diffusion trajectory. In high-efficiency frameworks, ill-timed injection can impede denoising and cause performance degradation. To address this, the LSR module employs a temporal scheduling coefficient $\gamma_{str}(t)$ to adaptively modulate guidance intensity, where $t\in\{0,\ldots,T-1\}$ is the discrete forward diffusion timestep $\gamma_{str}(t)$. The same timestep index $t$ is used in both the forward perturbation and the reverse sampling:

\begin{equation}
	\delta_{\mathrm{str}} = \gamma_{\mathrm{str}}(t)\delta_{norm}. 
\end{equation}
During reverse sampling, the timestep is traversed from $T-1$ to $0$. Accordingly, the structural guidance is strongest at the initial sampling steps and gradually decays to zero as the latent converges to the clean HR state.

The LSR module is consistently integrated into both the forward and reverse processes. During training, the structural prior $\delta_{\mathrm{str}}$ is incorporated into the residual shift, so that the forward transition is explicitly biased toward structure-aware degradation trajectories. The modified forward distribution can be written as:
\begin{equation}
	q(z_t \mid z_0, z_y, \delta_{\mathrm{str}})
	=
	\mathcal{N}\left(
	z_t;\,
	z_0 + \alpha_t \bigl(e_0 + \delta_{\mathrm{str}}\bigr),\,
	\kappa^2 \alpha_t \mathbf{I}
	\right).
\end{equation}
Equivalently, the latent state at time step $t$ admits the reparameterized form:
\begin{equation}
	z_t
	=
	z_0 + \alpha_t \bigl(e_0 + \delta_{\mathrm{str}}\bigr)
	+
	\kappa \sqrt{\alpha_t}\,\epsilon,
	\qquad
	\epsilon \sim \mathcal{N}(0,\mathbf{I}).
\end{equation}
In this way, the denoising model is encouraged to exploit the injected structural priors as compensation cues during reconstruction, thereby enhancing the recovery of local structures.

During reverse sampling, the structural prior $\delta_{\mathrm{str}}$ is injected into the intermediate latent state at each step:
\begin{equation}
	\tilde z_t=z_t+\alpha_t\delta_{\mathrm{str}}(t).
\end{equation}
The augmented latent representation is then processed by the denoising backbone, so that each iterative latent prediction remains guided by explicit local geometric constraints.

Although implemented as an on-the-fly modulation of the latent state $z_t$, this intervention is consistent with the structural correction term introduced in Eq. \ref{all}. Such a symmetric design improves the distributional alignment between training and inference. As a result, the structural prior helps suppress the tendency of the generation process to drift toward the visible-light manifold, while allowing the GRM module to focus on global modal mapping and statistical compensation. The combination of the two modules improves both optimization stability and reconstruction fidelity.

\section{Experiments}

\begin{figure*}[h]
	\centering
	\includegraphics[width=\textwidth]{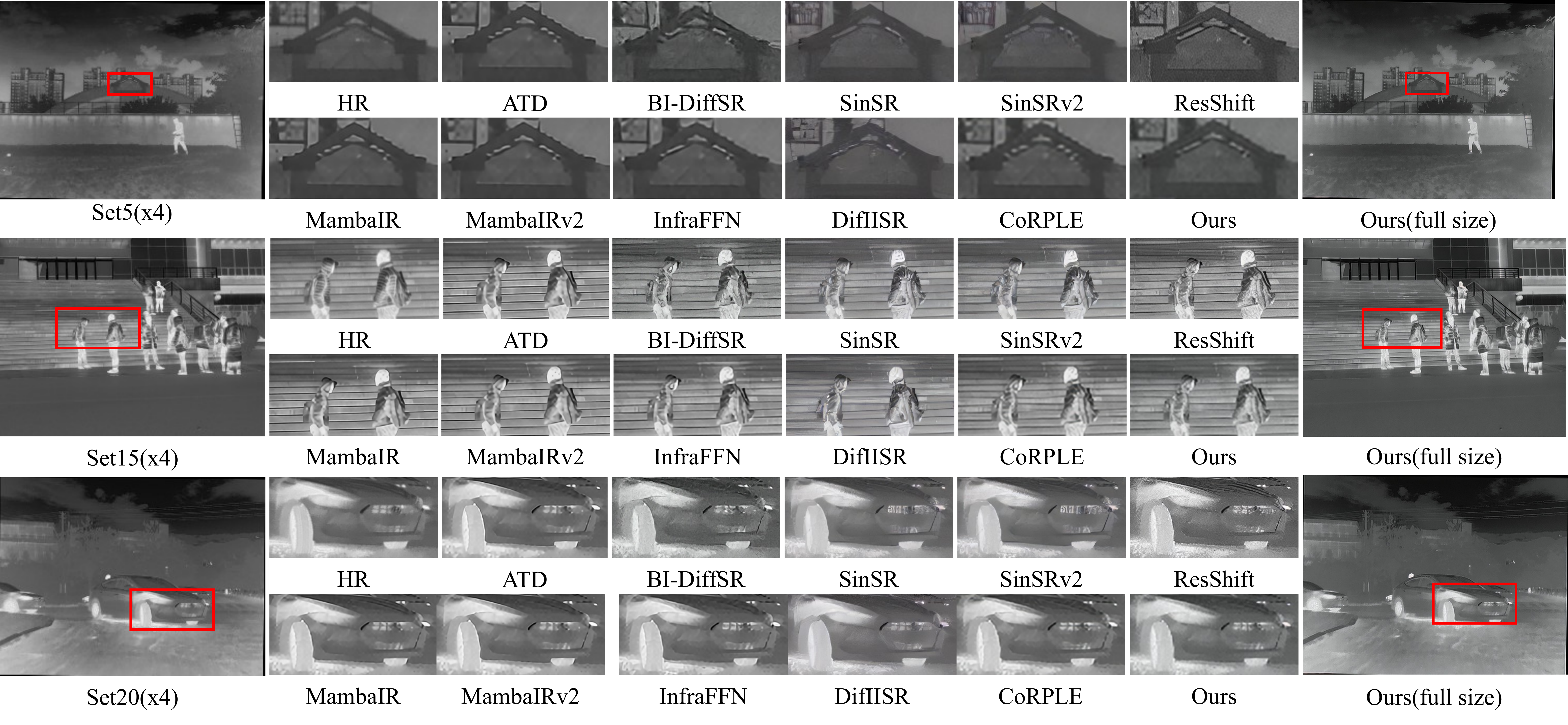}
	\caption{Qualitative comparison of different super-resolution methods on the M$^{3}$FD datasets.}
	\label{fig:qualitative_comparison}
\end{figure*}

\begin{figure*}[h]
	\centering
	\includegraphics[width=\textwidth]{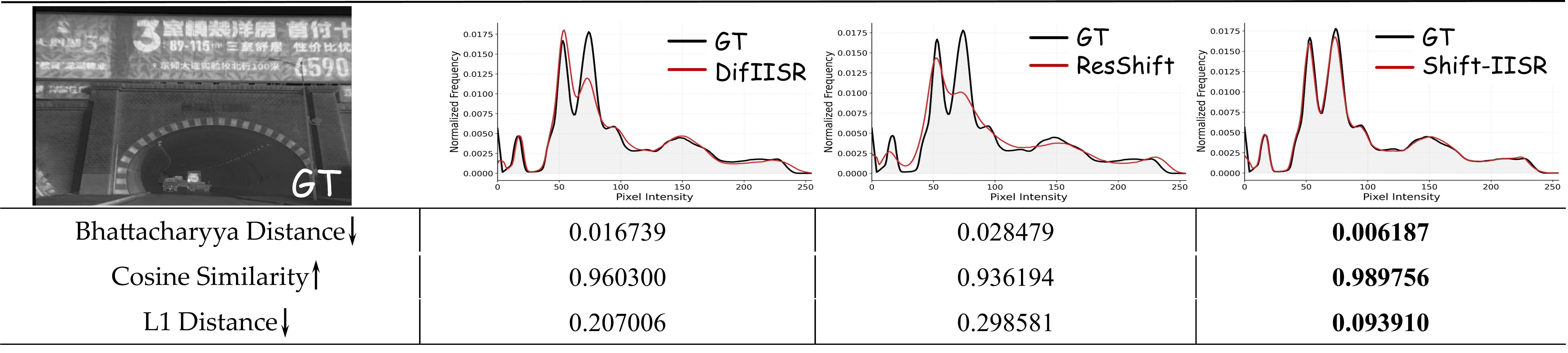}
	\caption{Comparison of Grayscale Histogram Distributions Between Reconstructed Images and Ground Truth.}
	\label{fig:distribution}
\end{figure*}
\subsection{Experimental Settings}

\textbf{Datasets and Evaluation Metrics.} We train our model on M$^{3}$FD \cite{liu2022target} and evaluate it on M$^{3}$FD \cite{liu2022target}, RoadScene \cite{xu2020u2fusion}, and TNO \cite{toet2017tno}. Following prior works such as CoRPLE \cite{li2024contourlet} and DifIISR \cite{li2025difiisr}, the M$^{3}$FD test set is further divided into three subsets: Set5, Set15, and Set20, to ensure fair comparison. We adopt three widely used full-reference metrics, including PSNR, SSIM \cite{wang2004image}, and LPIPS \cite{zhang2018unreasonable}, to assess reconstruction fidelity, structural similarity, and perceptual quality, respectively. In addition, we evaluate the distributional consistency between the super-resolved results and the ground truth using Cosine Similarity \cite{salton1975vector}, Bhattacharyya Distance \cite{bhattacharyya1943measure}, and L1 Distance. To further assess the practical value of reconstructed infrared images, we conduct downstream object detection and semantic segmentation experiments using YOLOv5s \cite{redmon2016you} on M$^{3}$FD and DeepLabv3+ \cite{chen2018encoder} on MSRS \cite{Tang2024Mask-DiFuser}.

\textbf{Experiment Details.} Shift-IISR was implemented and trained on a single NVIDIA GeForce RTX 4090 GPU. We adopted ResShift \cite{yue2023resshift}, a well-established framework in generative super-resolution, as our core backbone. Its underlying residual learning paradigm provides a natural architectural synergy with our proposed LSR module. Notably, we utilized the original pre-trained backbone initialization without any infrared-specific pre-training. The model converges within 20k iterations, achieving competitive performance across all benchmarks.

\subsection{Comparisons to State-of-the-Art Methods}

Extensive comparisons are conducted between Shift-IISR and existing mainstream super-resolution methods. The comparison covers both RGB SR models, including ResShift \cite{yue2023resshift}, SinSR and its variant SinSRv2 \cite{wang2024sinsr}, BI-DiffSR \cite{chen2024binarized}, ATD \cite{zhang2024transcending}, MambaIR \cite{guo2024mambair}, MambaIRv2 \cite{guo2025mambairv2}, and IISR methods, such as CoRPLE \cite{li2024contourlet}, InfraFFN \cite{qin2025infraffn}, DifIISR \cite{li2025difiisr}. Quantitative evaluation metrics and qualitative visual reconstruction results are respectively presented in Tab. \ref{tab:quantitative_comparison}, Fig. \ref{fig:qualitative_comparison} and  Fig. \ref{fig:distribution}. Only partial results are shown here; more comprehensive results can be found in the supplementary material.

\textbf{Quantitative Comparison.} Tab. \ref{tab:quantitative_comparison} summarizes the quantitative results of all compared methods. Shift-IISR achieves competitive overall performance, obtaining the best SSIM and LPIPS, which demonstrates its advantage in preserving structural information and perceptual fidelity. The comparison also reveals a clear domain gap between general-purpose and infrared-specific super-resolution models, as methods trained on visible images struggle to adapt to the thermal characteristics of infrared data. In addition, although DifIISR is fine-tuned on infrared images based on ResShift, its performance remains limited and noticeable color distortion artifacts are still observed.

\textbf{Qualitative Comparison.} Fig.~\ref{fig:qualitative_comparison} compares Shift-IISR with representative methods on samples from three subsets of M$^{3}$FD. Our method produces HR results that are closer to the ground truth, with improved visual quality and structural fidelity in narrow and complex regions. Despite its strong quantitative performance, CoRPLE still yields slightly blurry results and misses fine structural details. We further compare normalized grayscale histograms with the infrared ground truth in Fig.~\ref{fig:distribution}. Shift-IISR achieves better distributional consistency than ResShift and DifIISR, as supported by Cosine Similarity, Bhattacharyya Distance, and L1 Distance. Additional feature-space analysis is provided in the supplementary material.

\subsection{Ablation Study}
We performed an ablation study on the M$^{3}$FD dataset to validate the individual and collective contributions of our core components. Specifically, we analyzed four distinct settings to isolate the impact of the LSR and GRM modules: a backbone configuration lacking both modules, two intermediate versions each incorporating a single module, and the final optimized model. This multi-faceted evaluation, comprising pixel-wise fidelity metrics and subjective visual analysis, demonstrates how each module addresses specific challenges in IISR. A sensitivity analysis of the key hyperparameters is provided in the
supplementary material.

\textbf{Quantitative Comparison.} As summarized in Tab. \ref{tab:ablation}, the experimental results show that both modules contribute positively to the final performance for high-quality IISR. While each component independently enhances the performance, the most significant performance gains are achieved when they are integrated, suggesting a synergistic relationship. Specifically, the consistent improvements in SSIM validate the efficacy of our LSR module in preserving structural integrity and suppressing geometric distortions. Furthermore, the significant reduction in LPIPS underscores the efficacy of the GRM module in recovering realistic high-frequency details and enhancing overall perceptual fidelity.

\begin{table}[h]
	\centering
	\caption{Ablation study on the effectiveness of LSR and GRM.}
	\label{tab:ablation}
	\begin{tabular}{@{\extracolsep{\fill}}ccccc@{}}
		\toprule
		LSR Module & GRM Module & PSNR$\uparrow$ & LPIPS$\downarrow$ & SSIM$\uparrow$ \\
		\midrule
		-- & -- & 28.302 & 0.434 & 0.699 \\
		-- & $\checkmark$ & 30.718 & 0.328 & 0.815 \\
		$\checkmark$ & -- & 33.893 & 0.212 & 0.925 \\
		$\checkmark$ & $\checkmark$ & \textbf{34.099} & \textbf{0.174} & \textbf{0.931} \\
		\bottomrule
	\end{tabular}
\end{table}

\begin{figure}[!h]
	\centering
	\includegraphics[width=\linewidth]{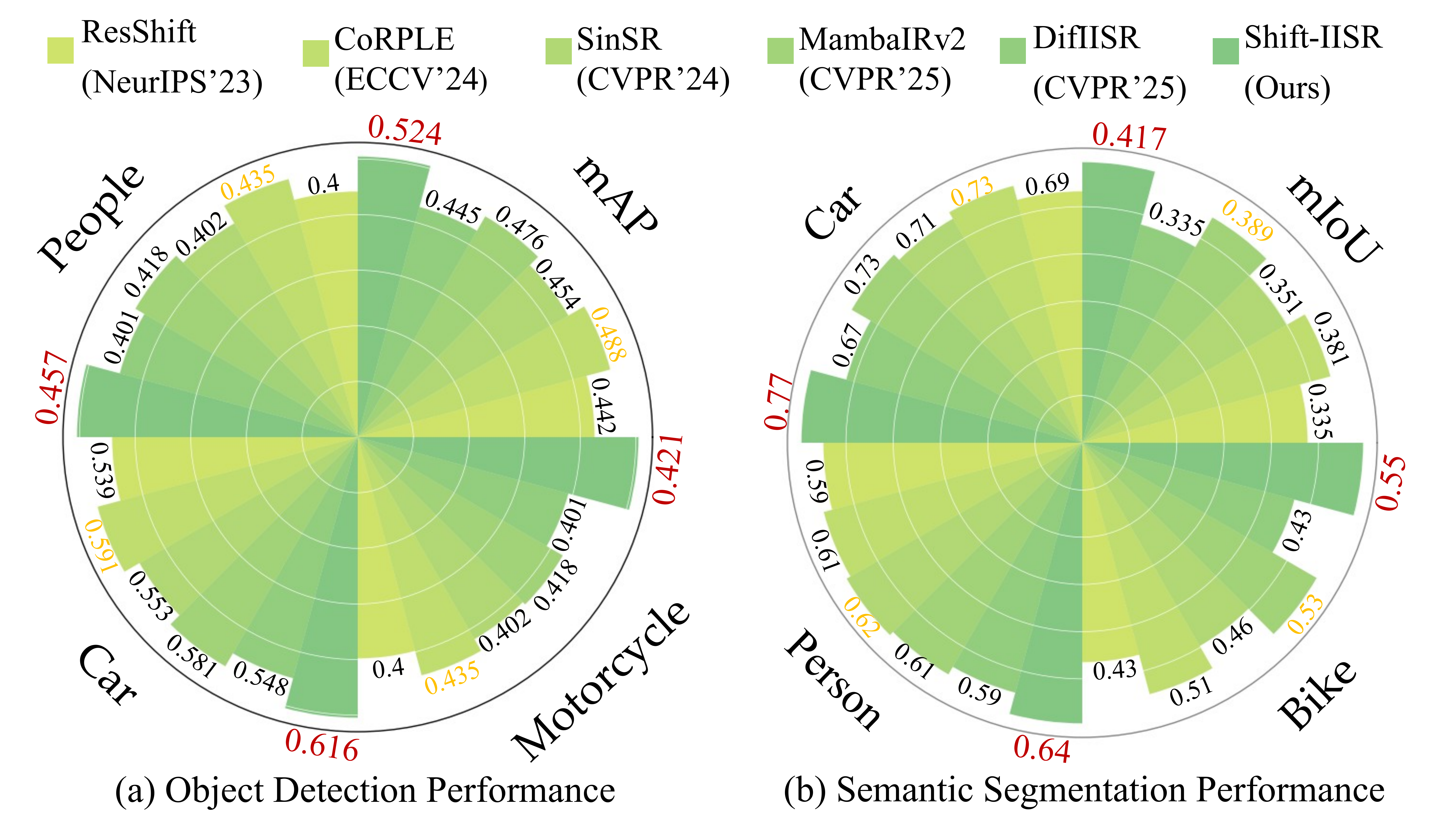}
	\caption{Quantitative results of different SR models on downstream applications for object detection and semantic segmentation.}
	\label{fig:DetAndSeg}
\end{figure}

\begin{figure*}[h]
	\centering
	\includegraphics[width=\textwidth]{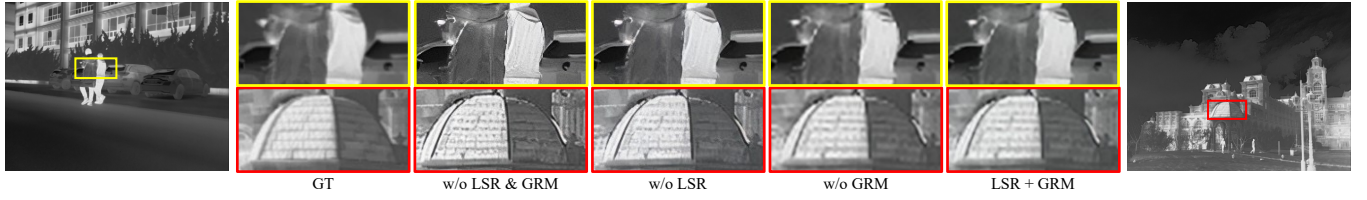}
	\caption{Visual comparison of generated results under different ablation settings.}
	\label{fig:ablation}
\end{figure*}

\textbf{Qualitative Comparison.} Fig.~\ref{fig:ablation} compares the contribution of each module through enlarged local patches. Removing LSR leads to clear structural degradation, such as missing details in the dome regions, demonstrating its importance for preserving geometric fidelity. The clothing textures further reveal a limitation of directly applying the RGB-oriented ResShift prior to IISR: it may generate modality-inconsistent hallucinated textures that deviate from the infrared manifold. Such cross-modality bias can produce less realistic local appearances. By injecting infrared-specific global representations, GRM effectively reduces the influence of these mismatched visible-spectrum priors and improves infrared-domain consistency.

\begin{figure*}[h]
	\centering
	\includegraphics[width=\textwidth]{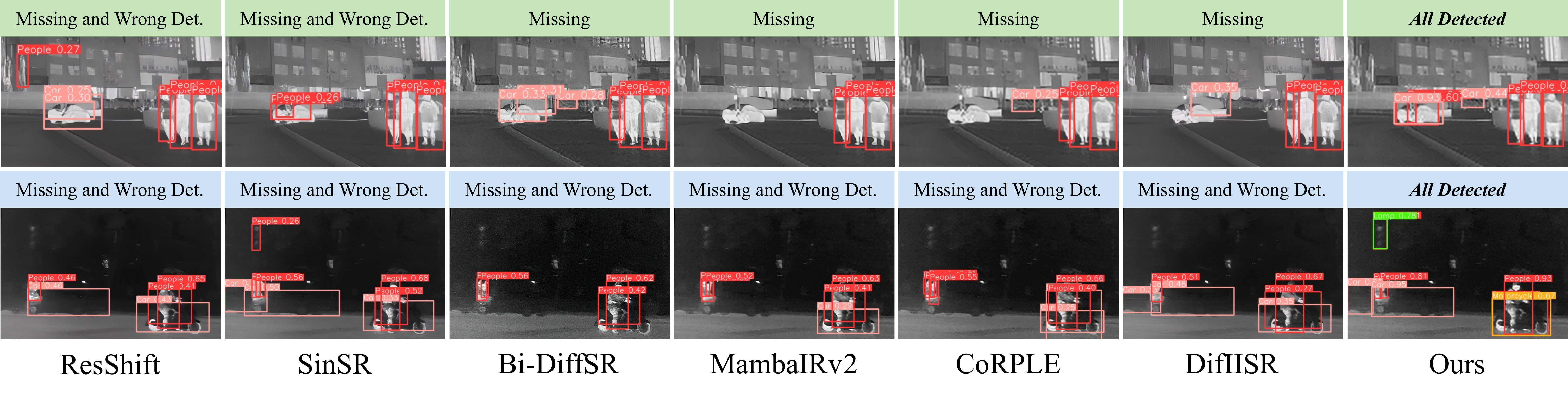}
	\caption{Comparison of Detection Performance for Infrared Image Super-Resolution Methods}
	\label{fig:Det}
\end{figure*}

\begin{figure*}[h]
	\centering
	\includegraphics[width=\textwidth]{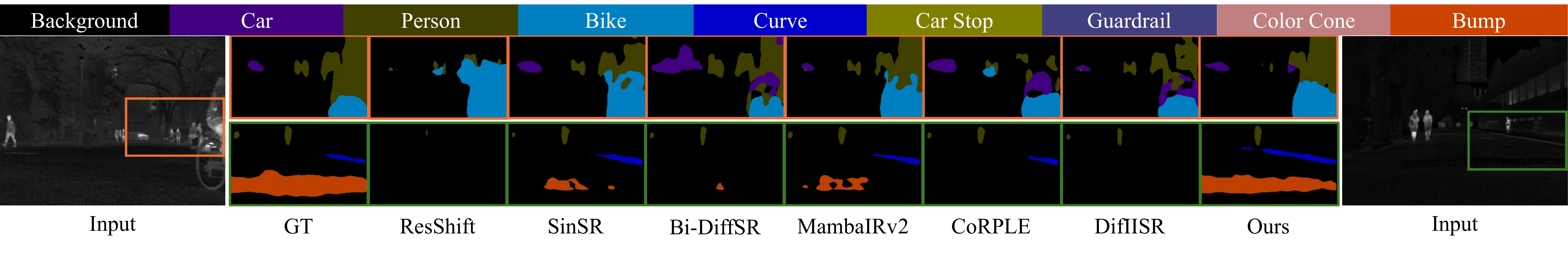}
	\caption{Comparison of Segmentation Performance for Infrared Image Super-Resolution Methods}
	\label{fig:Seg}
\end{figure*}

\subsection{Experiments on Infrared Object Detection}
\textbf{Setup.} For object detection, we adopt YOLOv5s \cite{redmon2016you}. To ensure fair comparison, the detector is first fine-tuned on the M$^3$FD training set and then evaluated on reconstructed images. The effectiveness of our SR framework is assessed through both quantitative mAP results and qualitative detection comparisons.

\textbf{Quantitative Comparison.} The left panel of Fig. \ref{fig:DetAndSeg} (a) illustrates the object detection performance evaluated on super-resolved images generated by various models. The results are reported both across the entire test set and within individual major categories. It is evident that our Shift-IISR consistently achieves superior detection accuracy, outperforming competing methods across all evaluated metrics and categories. This performance gain suggests that our model excels at restoring essential semantic features that are critical for downstream task recognition.

\textbf{Qualitative Comparison.} We provide a comparative visualization of detection results in Fig. \ref{fig:Det}. While alternative models frequently suffer from missing and wrong detection due to blurred boundaries and lost details, our model exhibits a robust capability to reliably identify and localize all targets.

\subsection{Experiments on Infrared Segmentation}

\textbf{Setup.} To evaluate the practical utility of the super-resolved images, we conducted semantic segmentation experiments using the DeepLabv3+ \cite{chen2018encoder} architecture. The segmentation model was fine-tuned on the MSRS \cite{Tang2024Mask-DiFuser} dataset  to ensure optimal feature alignment. Subsequently, we performed a comprehensive evaluation to compare the segmentation performance across different super-resolution models on the test set.

\textbf{Quantitative Comparison.} Quantitative results on downstream semantic segmentation are shown in the right panel of Fig. \ref{fig:DetAndSeg}(b). Our method achieves consistently better mIoU than the compared baselines, both overall and across individual classes, demonstrating the effectiveness of our SR framework for infrared semantic segmentation.

\textbf{Qualitative Comparison.} Representative segmentation results are shown in Fig. \ref{fig:Seg} to demonstrate the SR advantages of our method. Compared with other methods, our model produces more accurate masks with clearer boundaries, better semantic consistency, and improved preservation of fine structures, especially for small, thin, and overlapping objects.

\section{Conclusion}
In this paper, we address the fundamental mismatch between visible-oriented diffusion priors and the intrinsic demands of infrared image super-resolution. To tackle this issue, we propose Shift-IISR, a diffusion-based framework that decomposes the cross-modal discrepancy into two key aspects, namely global distribution shift and local structural heterogeneity, and progressively rectifies the generation process through GRM and LSR. Specifically, GRM is designed to recalibrate modality-related global representations, while LSR focuses on refining local structures to suppress visually plausible yet infrared-inconsistent details. Through the joint action of these two modules, the proposed method progressively steers the diffusion trajectory from the visible manifold toward the infrared manifold. Extensive experiments demonstrate that Shift-IISR effectively improves reconstruction quality while better preserving modality consistency across multiple benchmarks.

\bibliographystyle{ACM-Reference-Format}
\bibliography{sample-base}


\end{document}